%% file: obbvi-arxiv.tex
\documentclass[]{article}
\usepackage{proceed2e}

\input{preamble}

\input{preamble_acronyms}

\input{preamble_math}

\title{Overdispersed Black-Box Variational Inference}

\author{ {\bf Francisco J. R. Ruiz} \\
Columbia University\\
New York, USA \\
\And
{\bf Michalis K. Titsias}  \\
Athens University of Economics and Business\\
Athens, Greece \\
\And
{\bf David M. Blei}   \\
Columbia University\\
New York, USA \\
}

\begin{document}

\maketitle

\begin{abstract}

  We introduce overdispersed black-box variational inference, 
  a method to reduce the variance of the Monte Carlo
  estimator of the gradient in black-box variational inference.
  Instead of taking samples from the variational distribution, we use
  importance sampling to take samples from an overdispersed
  distribution in the same exponential family as the variational
  approximation. Our approach is general since it can be readily
  applied to any exponential family distribution, which is the typical
  choice for the variational approximation. We run experiments
  on two non-conjugate probabilistic models to show
  that our method effectively reduces the variance, and the
  overhead introduced by the computation of the proposal parameters
  and the importance weights is negligible. We find that
  our overdispersed importance sampling scheme provides lower variance
  than black-box variational inference, even when the latter uses twice
  the number of samples. This results in faster convergence of the
  black-box inference procedure.

\end{abstract}

\section{Introduction}

Generative probabilistic modeling is an effective approach for
understanding real-world data in many areas of science
\citep{Bishop2006,Murphy2012}. A probabilistic model describes a data-generating
process through a joint distribution of observed data and latent
(unobserved) variables. With a model in place, the investigator uses
an inference algorithm to calculate or approximate the posterior,
i.e., the conditional distribution of the latent variables given the
available observations. It is through the posterior that the
investigator explores the latent structure in the data and forms a
predictive distribution of future data. Approximating the posterior is
the central algorithmic problem for probabilistic modeling.

One of the most widely used methods to approximate the posterior
distribution is variational inference
\citep{Wainwright2008,Jordan1999}. Variational inference aims to
approximate the posterior with a simpler distribution, fitting that
distribution to be close to the exact posterior, where closeness is
measured in terms of \gls{KL} divergence. In minimizing the \gls{KL},
variational inference converts the problem of approximating the
posterior into an optimization problem.



Traditional variational inference uses coordinate ascent to optimize
its objective.  This works well for models in which each conditional
distribution is easy to compute~\citep{Ghahramani2001}, but is
difficult to use in more complex models where the variational
objective involves intractable expectations.  Recent
innovations in variational inference have addressed this with
stochastic optimization, forming noisy gradients with Monte Carlo
approximation. This strategy expands the scope of variational
inference beyond traditional models, e.g., to non-conjugate
probabilistic models
\citep{Carbonetto2009,Paisley2012,Salimans2013,Ranganath2014,Titsias2014_doubly},
deep neural
networks~\citep{Neal1992,Hinton1995,Mnih2014,Kingma2014,Ranganath2015},
and probabilistic programming~\citep{Wingate2013,Kucukelbir2015}. Some
of these techniques find their roots in classical policy search
algorithms for reinforcement learning
\citep{Williams1992,vandeMeent2016}.








These approaches must address a core problem with Monte Carlo
estimates of the gradient, which is that they suffer from high
variance. The estimated gradient can significantly differ from the
truth and this leads to slow convergence of the optimization. There
are several strategies to reduce the variance of the gradients,
including Rao-Blackwellization~\citep{Casella1996,Ranganath2014},
control variates~\citep{Ross2002,Paisley2012,Ranganath2014,Gu2016},
reparameterization
\citep{Price1958,Bonnet1964,Salimans2013,Kingma2014,Rezende2014,Kucukelbir2015},
and local expectations~\citep{Titsias2015}.

In this paper we develop \gls{OBBVI}, a new method for reducing the
variance of Monte Carlo gradients in variational inference.  The main
idea is to use importance sampling to estimate the gradient, in order
to construct a good proposal distribution that is matched to the
variational problem. We show that \gls{OBBVI} applies more
generally than methods such as reparameterization and local
expectations, and it further improves the profile of gradients that
use Rao-Blackwellization and control variates.

%
%

We demonstrate \gls{OBBVI} on two complex models: a non-conjugate
time series model~\citep{Ranganath2014} and Poisson-based \glspl{DEF}
\citep{Ranganath2015}. Our study shows that \gls{OBBVI} reduces the
variance of the original \gls{BBVI} estimates \citep{Ranganath2014},
even when using only half the number of Monte Carlo samples. This
provides significant savings in run-time complexity.

%


\parhead{Technical summary.} Consider a probabilistic model
$p(\bx,\bz)$, where $\bz$ are the latent variables and $\bx$ are the
observations. Variational inference sets up a parameterized
distribution of the latent variables $q(\bz; \blambda)$ and finds the
parameter $\blambda^\star$ that minimizes the \gls{KL} divergence
between $q(\bz; \blambda)$ and the posterior $p(\bz \g \bx)$. We then
use $q(\bz; \blambda^\star)$ as a proxy for the posterior.

We build on \gls{BBVI}, which solves this problem with a stochastic
optimization procedure that uses Monte Carlo estimates of the
gradient~\citep{Ranganath2014}. Let $\cL(\blambda)$ be the variational
objective, which is the (negative) \gls{KL}
divergence up to an additive constant. \gls{BBVI} uses samples from
$q(\bz; \blambda)$ to approximate its gradient,
\begin{align}\label{eq:reinforce}
	\nabla_{\blambda} \cL = \E{q(\bz; \blambda)}{f(\bz)},
\end{align}
where
\begin{align}\label{eq:fz}
  f(\bz) = \nabla_{\blambda} \log q(\bz;\blambda) \left(\log p(\bx,\bz)-\log q(\bz;\blambda)\right).
\end{align}
The resulting Monte Carlo estimator, based on sampling from
$q(\bz;\blambda)$, only requires evaluating the log-joint distribution
$\log p(\bz, \bx)$, the log-variational distribution
$\log q(\bz; \blambda)$, and the score function
$\nabla_{\blambda} \log q(\bz;\blambda)$.  Calculations about
$q(\bz; \blambda)$ can be derived once and stored in a library and, as
a consequence, \gls{BBVI} can be easily applied to a large class of
models. However, as we mentioned above, Monte Carlo estimates of this
gradient usually have high variance. \citet{Ranganath2014} correct for
this with Rao-Blackwellization and control variates.


%

We expand on this idea by approximating the gradient with importance
sampling.  We introduce a proposal distribution
$r(\bz; \blambda, \tau)$, which depends on both the variational
parameters and an additional parameter. (We discuss the additional
parameter below.) We then write the gradient as
\begin{align}
  \label{eq:isgrad}
  \nabla_{\blambda} \cL = \E{r(\bz; \blambda, \tau)}{f(\bz) \frac{q(\bz;
  \blambda)}{r(\bz; \blambda, \tau)}},
\end{align}
and form noisy estimates with samples from the proposal.

The key idea behind our method is that the optimal proposal
distribution (in terms of minimizing the variance of the resulting
estimator) is \textit{not} the original distribution
$q(\bz; \blambda)$~\citep[Chapter~9]{Owen2013}.  Rather, the optimal
proposal is a skewed version of that distribution with heavier tails.
Unfortunately, this distribution is not available to us---it involves
an intractable normalization constant.
But we use this insight to set a proposal with heavier tails than
the variational distribution, thus making it closer to the optimal
proposal. Note this is an unconventional use of importance
sampling, which is usually employed to approximate expectations of
intractable distributions. Here we use importance sampling to
improve the characteristics of a Monte Carlo estimator by sampling
from a different distribution.

In detail, we first assume that the variational distribution is in the
exponential family.  (This is not an assumption about the model; most
applications of variational inference use exponential family
variational distributions.)  We then set the proposal distribution to
be in the corresponding overdispersed exponential
family~\citep{Jorgensen1987}, where $\tau$ is the dispersion
parameter. We show that the corresponding estimator has lower
variance than the \gls{BBVI} estimator, we put forward a method to adapt
the dispersion parameter during optimization, and we demonstrate that this
method is more efficient than \gls{BBVI}. We call our approach
\emph{\acrlong{OBBVI}} (\acrshort{OBBVI}).



\parhead{Organization.} The rest of the paper is organized as
follows. We review \gls{BBVI} in Section~\ref{sec:bbvi}. We develop
\gls{OBBVI} in Section~\ref{sec:isbbvi}, describing both the basic
algorithm and its extensions to adaptive proposals and
high-dimensional settings.  Section~\ref{sec:experiments} reports on
our empirical study of two non-conjugate models.  We conclude the paper
in Section~\ref{sec:conclusions}.

\section{Black-Box Variational Inference}\label{sec:bbvi}

Consider a probabilistic model $p(\bx,\bz)$ and a variational family
$q(\bz ; \blambda)$ which is in the exponential family, i.e.,
\begin{equation}\label{eq:exp_fam}
  q(\bz;\blambda)=g(\bz)\exp\left\{ \blambda^\top t(\bz) - A(\blambda)\right\},
\end{equation}
where $g(\bz)$ is the base measure, $\blambda$ are the natural
parameters, $t(\bz)$ are the sufficient statistics, and $A(\blambda)$
is the log-normalizer. We are interested in a variational approximation
to the intractable posterior $p(\bz\g \bx)$, i.e., we aim to minimize
the \gls{KL} divergence $\KL{q(\bz;\blambda)}{p(\bz\g \bx)}$ with
respect to $\blambda$ \citep{Jordan1999}.  This is equivalent to
maximizing the \gls{ELBO},
\begin{equation}\label{eq:elbo}
  \Lcal(\blambda) = \E{q(\bz;\blambda)}{\log p(\bx,\bz)-\log q(\bz;\blambda)},
\end{equation}
which is a lower bound on the log of the marginal probability of the
observations, $\log p(\bx)$.

With a tractable variational family (e.g., the mean-field family) and a
conditionally conjugate model,\footnote{A conditionally conjugate
model is a model for which all the complete conditionals (i.e., the
posterior distribution of each hidden variable conditioned on the
observations and the rest of hidden variables) are in the same
exponential family as the prior.} the expectations in
Eq.~\ref{eq:elbo} can be computed in closed form and we can use
coordinate-ascent variational
inference~\citep{Ghahramani2001}. However, many models of interest
are not conditionally conjugate.  For these models, we need
alternative methods to optimize the \gls{ELBO}.  One approach is
\gls{BBVI}, which uses Monte Carlo estimates of the gradient and
requires few model-specific calculations~\citep{Ranganath2014}.
Thus, \gls{BBVI} is a variational inference algorithm that can be applied to
a large class of models.

\gls{BBVI} relies on the ``log-derivative trick,''
also called \textsc{reinforce} or score function method
\citep{Williams1992,Kleijnen1996,Glynn1990}, to obtain Monte
Carlo estimates of the gradient. In detail, we recover
the Monte Carlo estimate driven by Eqs.~\ref{eq:reinforce} and \ref{eq:fz}
by taking the gradient of \eqref{eq:elbo} with respect to the variational
parameters $\blambda$, and then applying the following two identities:
\begin{align}
  & \nabla_{\blambda} q(\bz;\blambda) =
                                      q(\bz;\blambda)\nabla_{\blambda}\log
                                      q(\bz;\blambda), \\
  & \E{q(\bz;\blambda)}{\nabla_{\blambda} \log q(\bz;\blambda)} =0 .
\end{align}
Eq.~\ref{eq:reinforce} enables noisy gradients of the \gls{ELBO} by
taking samples from $q(\bz;\blambda)$. However, the resulting
estimator may have high variance.  This is especially the case when
the variational distribution $q(\bz;\blambda)$ is a poor fit to the
posterior $p(\bz\g \bx)$, which is typical in early iterations of
optimization.




In order to reduce the variance of the estimator, \gls{BBVI} uses two
strategies: control variates and Rao-Blackwellization.  Because we
will also use these ideas in our algorithm, we briefly discuss them
here.

\parhead{Control variates.}  A control variate is a random variable
that is included in the estimator, preserving its expectation but
reducing its variance~\citep{Ross2002}. Although there are many
possible choices for control variates, \citet{Ranganath2014} advocate
for the weighted score function because it is not model-dependent.
Denote the score function by
$h(\bz) = \nabla_{\blambda} \log q(\bz;\blambda)$, and note again that
its expected value is zero.  With this function, each component $n$ of
the gradient in \eqref{eq:reinforce} can be rewritten as
$\E{q(\bz;\blambda)}{f_n(\bz)-a_n h_n(\bz)}$ where $a_n$ is a constant
and $f(\bz)$ is defined in \eqref{eq:fz}.  (Here, $f_n(\bz)$ and
$h_n(\bz)$ denote the $n$-th component of $f(\bz)$ and $h(\bz)$,
respectively.) We can set each
element $a_n$ to minimize the variance of the Monte Carlo estimates of
this expectation,
\begin{equation}\label{eq:a_n}
  a_n = \frac{\textrm{Cov}(f_n(\bz),h_n(\bz))}{\textrm{Var}(h_n(\bz))}.
\end{equation}
In \gls{BBVI}, a separate set of samples
from $q(\bz;\blambda)$ is used to estimate $a_n$ (otherwise, the
estimator would be biased).

%

%

\parhead{Rao-Blackwellization.}
Rao-Blackwellization~\citep{Casella1996} reduces the variance of a
random variable by replacing it with its conditional expectation,
given a subset of other variables. In \gls{BBVI}, each component of
the gradient is Rao-Blackwellized with respect to variables outside of
the Markov blanket of the involved hidden variable. More precisely,
assume a mean-field\footnote{A structured
  variational approach is also amenable to Rao-Blackwellization, but
  we assume a fully factorized variational distribution for
  simplicity.} variational distribution
$q(\bz;\blambda)=\prod_n q(z_n;\lambda_n)$.
We can equivalently rewrite the expectation of each
element in Eq.~\ref{eq:reinforce} as
\begin{align}
  \label{eq:rao-blackwell}
  \begin{split}
    \nabla_{\lambda_n} \Lcal = \mathbb{E}_{q(\bz_{(n)};\blambda_{(n)})} \big[& \nabla_{\lambda_n} \log q(z_n;\lambda_n) \\
    & \times\left(\log p_n(\bx,\bz_{(n)})-\log q(z_n;\lambda_n)\right) \big],
  \end{split}
\end{align}
where $\bz_{(n)}$ denotes the variable $z_n$ together with all latent
variables in its Markov blanket, $q(\bz_{(n)};\blambda_{(n)})$ denotes the
variational distribution on $\bz_{(n)}$, and $\log p_n(\bx,\bz_{(n)})$
contains all terms of the log-joint distribution that depend on
$\bz_{(n)}$.  The Monte Carlo estimate based on the Rao-Blackwellized
expectation has significantly smaller variance than the estimator
driven by Eq.~\ref{eq:reinforce}.

\section{Overdispersed Black-Box Variational Inference}
\label{sec:isbbvi}

We have described \gls{BBVI} and its two strategies for reducing the
variance of the noisy gradient.  We now describe \gls{OBBVI}, a method
for further reducing the variance.  The main idea is to use importance
sampling~\citep{Robert2005,Rubinstein2011} to estimate the gradient.
We first describe \gls{OBBVI} and the proposal distribution it uses.
We then show that this reduces variance, discuss several important
implementation details, and present the full algorithm.

\gls{OBBVI} does not sample from the variational distribution
$q(\bz;\blambda)$ to estimate the expectation
$\E{q(\bz;\blambda)}{f(\bz)}$.  Rather, it takes samples from a
proposal distribution $r(\bz; \blambda, \tau)$ and constructs
estimates of the gradient in Eq.~\ref{eq:isgrad}, where the
importance weights are
$w(\bz) = q(\bz;\blambda)/r(\bz; \blambda, \tau)$. This guarantees
that the resulting estimator is unbiased. The proposal
distribution involves the current setting of the variational
parameters $\blambda$ and an additional parameter $\tau$. 

\parhead{The optimal proposal.} The particular proposal that
\gls{OBBVI} uses is inspired by a result from the importance sampling
literature~\citep{Robert2005,Owen2013}. This result
states that the optimal proposal distribution, which minimizes the
variance of the estimator, is \emph{not} the variational distribution
$q(\bz;\blambda)$. Rather, the optimal proposal is
\begin{align}
  \label{eq:optimal-proposal}
  r_n^{\star}(\bz)\propto q(\bz;\blambda) |f_n(\bz)|,
\end{align}
for each component $n$ of the gradient. (Recall that $f(\bz)$ is a
vector of the same length as $\blambda$.)

%


While interesting, the optimal proposal distribution is not tractable
in general---it involves normalizing a complex product---and is not
``black box'' in the sense that it depends on the model
via $f(\bz)$. In \gls{OBBVI}, we build an
alternative proposal based on overdispersed exponential
families~\citep{Jorgensen1987}.  We will argue that this proposal is
closer to the (intractable) optimal $r^*(\bz)$ than the variational
distribution $q(\bz;\blambda)$, and that it is still practical in the
context of stochastic optimization of the variational objective.

\parhead{The overdispersed proposal.} Our motivation for using
overdispersed exponential families is that the optimal distribution of
Eq.~\ref{eq:optimal-proposal} assigns higher probability density to
the tails of $q(\bz;\blambda)$. There are two reasons for this
fact. First, consider settings of the variational parameters where the
variational distribution is a poor fit to the posterior.  For these
parameters, there are values of $\bz$ for which the posterior is high
but the variational distribution is small. While the optimal proposal
would sample configurations of $\bz$ for which $f_n(\bz)$ is large,
these realizations are in the tails of the variational distribution.

%

The second reason has to do with the score function.  The score
function $h_n(\bz)$ vanishes for values of $\bz$ for which the $n$-th
sufficient statistic $t_n(\bz)$ equals its expected value, and this
pushes probability mass (in the optimal proposal) to the tails of
$q(\bz; \blambda)$. To see this, recall the exponential family form
of the variational distribution given in Eq.~\ref{eq:exp_fam}. For
any exponential family distribution, the score function
is $h_n(\bz)=t_n(\bz)-\E{q(\bz;\blambda)}{t_n(\bz)}$. (This result
follows from simple properties of exponential families.\footnote{The
  gradient of the log-normalizer (with respect to the natural
  parameters) equals the first-order moment of the sufficient
  statistics, i.e.,
  $\nabla_{\blambda}A(\blambda)=\E{q(\bz;\blambda)}{t(\bz)}$.}) For
values of $\bz$ for which $t_n(\bz)$ is close to its expectation,
$h_n(\bz)$ becomes very close to zero. This zeros out $f_n(\bz)$ in
Eq.~\ref{eq:optimal-proposal}, which pushes mass to other parts of
$q(\bz;\blambda)$. As an example, in the case where $q(\bz;\blambda)$ is a
Gaussian distribution, the optimal proposal distribution places zero
mass on the mean of that Gaussian and hence more probability mass on
its tails.





Thus, we design a proposal distribution $r(\bz; \blambda, \tau)$ that
assigns higher mass to the tails of $q(\bz;\blambda)$.  Specifically,
we use an overdispersed distribution in the same exponential family as
$q(\bz;\blambda)$. The proposal is
\begin{align}
  r(\bz;\blambda,\tau) = g(\bz,\tau)\exp\left\{ \frac{\blambda^\top t(\bz) - A(\blambda)}{\tau}\right\},
\end{align}
where $\tau\geq 1$ is the dispersion coefficient of the overdispersed
distribution~\citep{Jorgensen1987}. Hence, the \gls{OBBVI} estimator
of the gradient can be expressed as
\begin{equation}\label{eq:est_grad_obb}
    \widehat{\nabla}_\blambda^{\textrm{O-BB}} \Lcal = \frac{1}{S}\sum_s f(\bz^{(s)})\frac{q(\bz^{(s)};\blambda)}{r(\bz^{(s)})}, \quad \bz^{(s)} \stackrel{\textrm{iid}}{\sim} r(\bz;\blambda,\tau),
\end{equation}
where $S$ is the number of samples of the Monte Carlo approximation.

This choice of $r(\bz;\blambda,\tau)$ has several desired properties
for a proposal distribution.  First, it is easy to sample from, since
for fixed values of $\tau$ it belongs to the same exponential family as
$q(\bz;\blambda)$.  Second, as for the optimal proposal, it is
adaptive, since it explicitly depends on the parameters $\blambda$ which we
are optimizing. Finally, by definition, it assigns higher mass to the
tails of $q(\bz;\blambda)$, which was our motivation for choosing it.

The dispersion coefficient $\tau$ can be itself adaptive to better
match the optimal proposal at each iteration of the variational
optimization procedure. We put forward a method to update the value of
$\tau$ in Section~\ref{sec:extensions}.

Note that our approach differs from importance weighted autoencoders
\citep{Burda2016}, which also make use of importance sampling but with
the goal of deriving a tighter log-likelihood lower bound in the
context of the variational autoencoder \citep{Kingma2014}. In
contrast, we use importance sampling to reduce the variance of the
estimator of the gradient.

\subsection{Variance reduction}

Here, we compare the variance of the \gls{OBBVI} estimator
$\widehat{\nabla}^{\textrm{O-BB}}_\lambda \Lcal$ given in
Eq.~\ref{eq:est_grad_obb} with the variance of the original \gls{BBVI}
estimator $\widehat{\nabla}^{\textrm{BB}}_\lambda \Lcal$, which samples
from $q(\bz;\blambda)$:
\begin{align}
    \widehat{\nabla}_\blambda^{\textrm{BB}} \Lcal = \frac{1}{S}\sum_s f(\bz^{(s)}), \qquad \bz^{(s)}\stackrel{\textrm{iid}}{\sim} q(\bz;\blambda).
\end{align}
After some algebra, we can express the variance of the \gls{BBVI} estimator as
\begin{equation}
    \Var{\widehat{\nabla}^{\textrm{BB}}_\blambda \Lcal} = \frac{1}{S}\E{q(\bz;\blambda)}{f^2(\bz)}-\frac{1}{S}(\nabla_\blambda\Lcal)^2,
\end{equation}
and we can also express the variance of the \gls{OBBVI} estimator in terms of an expectation with respect to the variational distribution as
\begin{equation}\label{eq:varL2}
  \begin{split}
    \Var{\widehat{\nabla}^{\textrm{O-BB}}_\blambda \Lcal} & = \frac{1}{S}\E{r(\bz;\blambda,\tau)}{f^2(\bz)\frac{q^2(\bz;\blambda)}{r^2(\bz;\blambda,\tau)}}-\frac{1}{S}(\nabla_\blambda\Lcal)^2\\
    & = \frac{1}{S}\E{q(\bz;\blambda)}{f^2(\bz)\frac{q(\bz;\blambda)}{r(\bz;\blambda,\tau)}}-\frac{1}{S}(\nabla_\blambda\Lcal)^2.
  \end{split}
\end{equation}
Variance reduction for the \gls{OBBVI} approach is achieved when $\Var{\widehat{\nabla}^{\textrm{O-BB}}_\blambda \Lcal}\leq \Var{\widehat{\nabla}^{\textrm{BB}}_\blambda \Lcal}$ or, equivalently,
\begin{equation}\label{eq:varianceReduction}
  \E{q(\bz;\blambda)}{f^2(\bz)\frac{q(\bz;\blambda)}{r(\bz;\blambda,\tau)}}\leq \E{q(\bz;\blambda)}{f^2(\bz)}.
\end{equation}
This inequality  is trivially satisfied when we set $r(\bz)$ to the optimal proposal distribution, $r^{\star}(\bz)$. Moreover, Eq.~\ref{eq:varianceReduction} also gives us some intuition on why the use of an overdispersed proposal distribution can reduce the variance, since $r(\bz;\blambda,\tau)$ will be larger than $q(\bz;\blambda)$ for those values of $\bz$ for which the product $q(\bz;\blambda)f^2(\bz)$ is highest, i.e., in the tails of $q(\bz;\blambda)$. Our experimental results in Section~\ref{sec:experiments} demonstrate that the variance is effectively reduced when we use our \gls{OBBVI}.



\subsection{Implementation}\label{sec:extensions}

We now discuss several extensions of \gls{OBBVI} that make it more
suitable for real applications.

\paragraph{High dimensionality.}
Previously, we defined the proposal distribution $r(\bz;\blambda,\tau)$ as an overdispersed version of the variational distribution $q(\bz;\blambda)$. However, importance sampling is known to fail when the dimensionality of the hidden space is moderately high, due to the high resulting variance of the importance weights $w(\bz)=q(\bz;\blambda)/r(\bz;\blambda,\tau)$. To address this, we rely on the fact that hidden variable $z_n$ is the variable with the highest influence on the estimator of the $n$-th component of the gradient. We exploit this idea, which was also considered by \citet{Titsias2015} in their algorithm based on local expectations.


More precisely, for the variational parameters of variable $z_n$, we first write the gradient as
\begin{equation}\label{eq:isbbvi_high_dim}
	\begin{split}
		\nabla_{\lambda_n}\Lcal & = \E{q(z_n;\lambda_n)}{\E{q(\bz_{\neg n};\blambda_{\neg n})}{f_n(\bz)}}\\
		& = \E{r(z_n;\lambda_n,\tau_n)}{w(z_n)\E{q(\bz_{\neg n};\blambda_{\neg n})}{f_n(\bz)}},
	\end{split}
\end{equation}
where $r(z_n;\lambda_n,\tau_n)$ is the overdispersed version of $q(z_n;\lambda_n)$ with dispersion coefficient $\tau_n$, $\bz_{\neg n}$ denotes all hidden variables in the model except $z_n$, and similarly for $\blambda_{\neg n}$. Thus, the corresponding importance weights in \eqref{eq:isbbvi_high_dim} for each component of the gradient depend only on variable $z_n$, i.e.,
\begin{equation}
	w(z_n) = \frac{q(z_n;\lambda_n)}{r(z_n;\lambda_n,\tau_n)}.
\end{equation}

We use a single sample from $q(\bz_{\neg n};\blambda_{\neg n})$ to estimate the inner expectation in \eqref{eq:isbbvi_high_dim}, and $S$ samples of $z_n$ from $r(z_n;\lambda_n,\tau_n)$ to estimate the outer expectation.

\paragraph{Adaptation of the dispersion coefficients.}
Our algorithm requires setting the value of the dispersion parameters $\tau_n$; we would like to automate this procedure. Here, we develop a method to learn these coefficients during  optimization by minimizing the variance of the estimator. More precisely, we introduce stochastic gradient descent steps for $\tau_{n}$ that minimize the variance. The exact derivative of the (negative) variance with respect to $\tau_{n}$ is
\begin{align}\label{eq:dvar_dtau}
	-\frac{\partial \Var{\widehat{\nabla}^{\textrm{O-BB}}_{\lambda_n} \Lcal}}{\partial \tau_n} 
	= & \frac{1}{S} \mathbb{E}_{r(z_n;\lambda_n,\tau_n)} \Bigg[ {\E{q(\bz_{\neg n};\blambda_{\neg n})}{f_n(\bz)}^2} \nonumber \\
	& \times w^2(z_n){\frac{\partial \log r(z_n;\lambda_n,\tau_n)}{\partial \tau_n}}\Bigg],
\end{align}
where we have applied the log-derivative trick once again, as well as the extension to high dimensionality detailed above. Now a Monte Carlo estimate of this derivative can be obtained by using the same set of $S$ samples used in the update of $\lambda_n$. The resulting procedure is fast, with little extra overhead, since both $f_n(\bz)$ and $w(z_n)$ have been pre-computed.

Thus, we perform gradient steps of the form
\begin{equation}\label{eq:adapttau}
  \tau^{(t)}_{n}=\tau^{(t-1)}_{n}- \alpha_n \frac{\partial \Var{\widehat{\nabla}^{\textrm{O-BB}}_{\lambda_n}\Lcal}}{\partial \tau_{n}},
\end{equation}
where $\tau_n$ is constrained as $\tau_n\geq 1$ and the derivatives are estimated via Monte Carlo approximation. Since the derivatives in Eq.~\ref{eq:adapttau} can be several orders of magnitude greater than $\tau_{n}$, we opt for a simple approach to choose an appropriate step size $\alpha_n$. In particular, we ignore the magnitude of the derivative in~\eqref{eq:adapttau} and take a small gradient step in the direction given by its sign. Note that we do not need to satisfy the Robbins-Monro conditions here \citep{Robbins1951}, because the adaptation of $\tau_n$ only defines the proposal distribution and it is not part of the original stochastic optimization procedure.

Eq.~\ref{eq:adapttau} can still be applied even if $\lambda_n$ is a vector; it only requires replacing the derivative of the variance with the summation of the derivatives for all components of $\lambda_n$.

\paragraph{Multiple importance sampling.}
It may be more stable (in terms of the variance of the importance weights) to consider a set of $J$ dispersion coefficients, $\tau_{n1},\ldots,\tau_{nJ}$, instead of a single coefficient $\tau_n$. We propose to use a mixture with equal weights to build the proposal as follows:
\begin{equation}
	r(z_n;\lambda_n,\tau_{n1},\ldots,\tau_{nJ}) = \frac{1}{J}\sum_{j=1}^{J} r(z_n;\lambda_n,\tau_{nj}),
\end{equation}
where each term in the mixture is given by $r(z_n;\lambda_n,\tau_{nj})=g(z_n,\tau_{nj})\exp\left\{ \frac{\lambda_n^\top t(z_n)-A(\lambda_n)}{\tau_{nj}}\right\}$. In the importance sampling literature, this is known as \gls{MIS}, as multiple proposals are used \citep{Veach1995}. Within the \gls{MIS} methods, we opt for full \gls{DMIS} because it is the approach that presents lowest variance \citep{Hesterberg1995,Owen2000,Elvira2015_generalized}. In \gls{DMIS}, the number of samples $S$ of the Monte Carlo estimator must be an integer multiple of the number of mixture components $J$, and $S/J$ samples are deterministically assigned to each proposal $r(z_n;\lambda_n,\tau_{nj})$. However, the importance weights are obtained as if the samples had been actually drawn from the mixture, i.e.,
\begin{equation}\label{eq:weights_dmis}
	w(z_n) = \frac{q(z_n;\lambda_n)}{\frac{1}{J}\sum_{j=1}^{J} r(z_n;\lambda_n,\tau_{nj})}.
\end{equation}
This choice of the importance weights yields an unbiased estimator with smaller variance than the standard \gls{MIS} approach \citep{Owen2000,Elvira2015_generalized}.

In the experiments in Section~\ref{sec:experiments} we investigate the performance of two-component proposal distributions, where  $J=2$, and compare it against our initial algorithm that uses a unique proposal, which corresponds to $J=1$. We have also conducted some additional experiments (not shown in the paper) with mixtures with higher number of components, with no significant improvements.

\subsection{Full algorithm}
We now present our full algorithm for \gls{OBBVI}. It makes use of control variates, Rao-Blackwellization, and overdispersed importance sampling with adaptation of the dispersion coefficients. At each iteration, we draw a single sample $\bz^{(0)}$ from the variational distribution, as well as $S$ samples $z_n^{(s)}$ from the overdispersed proposal for each $n$ (using \gls{DMIS} in this step). We obtain the score function as
\begin{equation}\label{eq:fullalg_hd}
  h_n(z_n^{(s)}) = \nabla_{\lambda_n} \log q(z_n^{(s)};\lambda_n),
\end{equation}
and the argument of the expectation in \eqref{eq:rao-blackwell} as
\begin{equation}\label{eq:fullalg_fd}
  f_n(\bz^{(s)})=h_n(z_n^{(s)})(\log p_n(\bx,z_n^{(s)},\bz_{\neg n}^{(0)})-\log q(z_n^{(s)};\lambda_n)),
\end{equation}
where $p_n$ indicates that we use Rao-Blackwellization. Finally, the estimator of the gradient is obtained as 
\begin{equation}\label{eq:fullalg_grad}
  \widehat{\nabla}_{\lambda_n}\Lcal = \frac{1}{S}\sum_s \left(f^{\textrm{w}}_n(\bz^{(s)})-a_n h_n^{\textrm{w}}(z_n^{(s)})\right),
\end{equation}
where the superscript `${\textrm{w}}$' stands for `weighted,' i.e.,
\begin{align}
  f^{\textrm{w}}_n(\bz^{(s)})=w(z_n^{(s)})f_n(\bz^{(s)}),\label{eq:fw}\\
  h^{\textrm{w}}_n(z_n^{(s)})=w(z_n^{(s)})h_n(z_n^{(s)}).\label{eq:hw}
\end{align}
Following Eq.~\ref{eq:a_n}, we use a separate set of samples to estimate the optimal $a_n$ as
\begin{equation}\label{eq:fullalg_optimal_ad}
  a_n=\frac{\widehat{\textrm{Cov}}(f^{\textrm{w}}_n,h^{\textrm{w}}_n)}{\widehat{\textrm{Var}}(h^{\textrm{w}}_n)}.
\end{equation}

We use AdaGrad \citep{Duchi2011} to obtain adaptive learning rates that ensure convergence of the stochastic optimization procedure, although other schedules can be used instead as long as they satisfy the standard Robbins-Monro conditions \citep{Robbins1951}. In AdaGrad, the learning rate is obtained as
\begin{equation}\label{eq:fullalg_rho_t}
	\brho_t = \eta\;\diag{\bG_t}^{-1/2},
\end{equation}
where $\bG_t$ is a matrix that contains the sum across the first $t$ iterations of the outer products of the gradient, and $\eta$ is a constant. Thus, the stochastic gradient step is given by
\begin{equation}\label{eq:fullalg_lambdanew}
  \blambda^{(t)} = \blambda^{(t-1)}+\brho_t\circ\widehat{\nabla}_{\blambda}\Lcal,
\end{equation}
where `$\circ$' denotes the element-wise (Hadamard) product. Algorithm~\ref{alg:isBBVI} summarizes the full procedure.

\begin{algorithm}[t]
    \SetAlgoLined
    \SetKwInOut{KwInput}{input}
    \SetKwInOut{KwOutput}{output}
    \KwInput{data $\bx$, joint distribution $p(\bx,\bz)$, mean-field variational family $q(\bz;\blambda)$}
    \KwOutput{variational parameters $\blambda$}
    Initialize $\blambda$\;
    Initialize the dispersion coefficients $\tau_{nj}$\;
    \While{algorithm has not converged}{
      \tcc{draw samples}
      Draw a single sample $\bz^{(0)}\sim q(\bz;\blambda)$\;
      \For{$n=1$ \emph{to} $N$}{
      Draw $S$ samples $z_n^{(s)}\sim r(z_n;\lambda_n,\{\tau_{nj}\})$ (\acrshort{DMIS})\;
	    Compute the importance weights $w(z_n^{(s)})$ (Eq.~\ref{eq:weights_dmis})\;
      }    
      \tcc{estimate gradient}
      \For{$n=1$ \emph{to} $N$}{
      For each sample $s$, compute $h_n(z_n^{(s)})$ (Eq.~\ref{eq:fullalg_hd})\;
      For each sample $s$, compute $f_n(\bz^{(s)})$ (Eq.~\ref{eq:fullalg_fd})\;
      Compute the weighted $f_n^{\textrm{w}}(\bz^{(s)})$ (Eq.~\ref{eq:fw})\;
      Compute the weighted $h_n^{\textrm{w}}(z_n^{(s)})$ (Eq.~\ref{eq:hw})\;
      Estimate the optimal $a_n$ (Eq.~\ref{eq:fullalg_optimal_ad})\;
      Estimate the gradient $\widehat{\nabla}_{\lambda_n}\Lcal$ (Eq.~\ref{eq:fullalg_grad})\;
    }        
    \tcc{update dispersion coefficients}
    \For{$n=1$ \emph{to} $N$}{
    	Estimate the derivatives $\frac{\partial \Var{\nabla_{\lambda_n}\Lcal}}{\partial \tau_{nj}}$ (Eq.~\ref{eq:dvar_dtau})\;
    	Take a gradient step for $\tau_{nj}$ (Eq.~\ref{eq:adapttau})\;
    }
    \tcc{take gradient step}
    Set the step size $\brho_t$ (Eq.~\ref{eq:fullalg_rho_t})\;
    Take a gradient step for $\blambda$ (Eq.~\ref{eq:fullalg_lambdanew})\;
    }
\caption{\Acrfull{OBBVI}\label{alg:isBBVI}}
\end{algorithm}

\input{sec_empirical_study}

\section{Conclusions}\label{sec:conclusions}
\glsresetall
We have developed \gls{OBBVI}, a method that relies on importance sampling
to reduce the variance of the stochastic gradients in \gls{BBVI}.
\gls{OBBVI} uses an importance sampling proposal distribution that has
heavier tails than the actual variational distribution.
In particular, we choose the proposal as an overdispersed distribution
in the same exponential family as the variational distribution.
Like \gls{BBVI}, our approach is amenable to
mean field or structured variational inference, as well as
variational models \citep{Ranganath2015_hvm,Tran2016}.

We have studied the performance of our method on two complex probabilistic models. Our results show that \gls{BBVI} effectively benefits from the use of overdispersed importance sampling, and \gls{OBBVI} leads to faster convergence in the resulting stochastic optimization procedure.

There are several avenues for future work. First, we can explore other proposal distributions to provide a better fit to the optimal ones while still maintaining computational efficiency. Second, we can apply quasi-Monte Carlo methods to further decrease the sampling variance, as already suggested by \citet{Ranganath2014}. Finally, we can combine the reparameterization trick with overdispersed proposals to explore whether variance is further reduced.




\bibliographystyle{apa}
\bibliography{obbvi-bib}


\end{document}

%% file: preamble.tex
\usepackage[T1]{fontenc}
\usepackage{tgtermes}
\usepackage[T1]{fontenc}
\usepackage{scalefnt,letltxmacro}
\LetLtxMacro{\oldtextsc}{\textsc}
\renewcommand{\textsc}[1]{\oldtextsc{\scalefont{1.10}#1}}

\usepackage{times,amsmath,amssymb,paralist,graphicx,subfig}
\usepackage[usenames,dvipsnames]{xcolor}
\usepackage[acronym,smallcaps,nowarn]{glossaries}
\usepackage{natbib}
\usepackage[algoruled]{algorithm2e}

\newcommand{\parhead}[1]{\textbf{#1}\,}

%% file: preamble_acronyms.tex
\newacronym{BBVI}{bbvi}{black-box variational inference}

\newacronym{CTM}{ctm}{correlated topic model}

\newacronym[\glslongpluralkey={deep exponential families}]{DEF}{def}{deep exponential family}
\newacronym{DMIS}{dmis}{deterministic multiple importance sampling}

\newacronym{ELBO}{elbo}{evidence lower bound}

\newacronym{GNTS}{gn-ts}{gamma-normal time series model}

\newacronym{KL}{kl}{{K}ullback-{L}eibler}

\newacronym{LDA}{lda}{latent {D}irichlet allocation}

\newacronym{MIS}{mis}{multiple importance sampling}

\newacronym{OBBVI}{o-bbvi}{overdispersed black-box variational inference}

\newacronym{SVI}{svi}{stochastic variational inference}

%% file: preamble_math.tex
\DeclareRobustCommand{\KL}[2]{\ensuremath{D_{\textrm{KL}}\left(#1\;\|\;#2\right)}}

\DeclareRobustCommand{\E}[2]{\mathbb{E}_{#1}\left[#2\right]}
\DeclareRobustCommand{\Var}[1]{\mathbb{V}\left[#1\right]}

\DeclareRobustCommand{\diag}[1]{\textrm{diag}\left(#1\right)}

\newcommand{\g}{\, | \,}

\newcommand{\Lcal}{\mathcal{L}}
\newcommand{\Ncal}{\mathcal{N}}

\newcommand{\bz}{\mathbf{z}}
\newcommand{\bx}{\mathbf{x}}
\newcommand{\bG}{\mathbf{G}}
\newcommand{\brho}{\mathbb{\rho}}
\newcommand{\blambda}{\mathbb{\lambda}}
\newcommand{\cL}{\mathcal{L}}

%% file: sec_empirical_study.tex
\section{Empirical Study}
\label{sec:experiments}
\glsresetall{}

We study our method with two non-conjugate probabilistic models: the
\gls{GNTS} and the Poisson \gls{DEF}. We found that \gls{OBBVI}
reduces the variance of the \gls{BBVI} estimator and leads to faster
convergence.

\subsection{Description of the experiments}
\label{sec:descr_experiments}

\parhead{Models description and datasets.}
The \gls{GNTS} model \citep{Ranganath2014} is a non-conjugate
state-space model for sequential data that was used to showcase
\gls{BBVI}. The model is described by
\begin{equation}
  \begin{split}
    & w_{kd}\sim \Ncal(0,\sigma_w^2),\\
    & o_{nd}\sim \Ncal(0,\sigma_o^2),\\
    & z_{n1k}\sim \textrm{GammaE}(\sigma_z,\sigma_z),\\
    & z_{ntk}\sim \textrm{GammaE}(z_{n(t-1)k},\sigma_z),\\
    & x_{ndt}\sim \Ncal\left(o_{nd}+\sum_k z_{ntk}w_{kd},\sigma_x^2\right).
  \end{split}
\end{equation}
The indices $n$, $t$, $d$ and $k$ denote observations, time instants,
observation dimensions, and latent factors, respectively. The
distribution $\textrm{GammaE}$ denotes the expectation/variance
parameterization of the gamma distribution.
The model explains each datapoint $x_{ndt}$ with a latent factor
model. For each time instant $t$, the mean of $x_{ndt}$ depends on the
inner product $\sum_k z_{ntk}w_{kd}$, where $z_{ntk}$ varies smoothly
across time. The variables $o_{nd}$ are an intercept that capture the
baseline in the observations.

We set the hyperparameters to be
$\sigma_w^2=1$, $\sigma_o^2=1$, $\sigma_z=1$, and $\sigma^2_x=0.01$.
We use a synthetic dataset of $N=900$ time sequences of length $T=30$
and dimensionality $D=20$. We use $K=30$ latent factors, leading to
$828,600$ hidden variables.

The Poisson \gls{DEF} \citep{Ranganath2015} is a multi-layered
latent variable model of discrete data, such as text. The model is
described by
\begin{equation}
  \begin{split}
    & w_{kv}^{(0)} \sim \textrm{Gamma}(\alpha_w,\beta_w),\\
    & w_{k k^\prime}^{(\ell)} \sim \textrm{Gamma}(\alpha_w,\beta_w),\\
    & z_{dk}^{(L)} \sim \textrm{Poisson}(\lambda_z),\\
    & z_{dk}^{(\ell)} \sim \textrm{Poisson}\left(\sum_{k^\prime} z_{dk^\prime}^{(\ell+1)} w_{k^\prime k}^{(\ell)}\right),\\
    & x_{dv} \sim \textrm{Poisson}\left(\sum_{k^\prime} z_{dk^\prime}^{(1)} w_{k^\prime v}^{(0)}\right).\\
  \end{split}
\end{equation}
The indices $d$, $v$, $k$ and $\ell$ denote documents, vocabulary
words, latent factors, and hidden layers, respectively. This model
captures a hierarchy of dependencies between latent variables similar
to the hidden structure in deep neural networks. In detail, the number
of times that word $v$ appears in document $d$ is $x_{dv}$. It has a
Poisson distribution with rate given by an inner product of
gamma-distributed weights and Poisson-distributed hidden variables
from layer $1$. The Poisson-distributed hidden variables depend, in
turn, on another set of weights and another layer of hidden
Poisson-distributed variables. This structure repeats for a specified
number of layers.

We set the prior shape and rate as $\alpha_w=0.1$ and $\beta_w=0.3$,
and the prior mean for the top level of the Poisson \gls{DEF} as
$\lambda_z=0.1$. We use $L=3$ layers with $K=50$ latent factors each.
We model the papers at the Neural Information Processing Systems
(NIPS) 2011 conference. This is a data set with $D=305$ documents,
$612,508$ words, and $V=5715$ vocabulary words (after removing stop
words). This leads to a model with $336,500$ hidden variables.


\parhead{Evaluation.} We compare \gls{OBBVI} with
\gls{BBVI}~\citep{Ranganath2014}. For a fair comparison, we use the
same number of samples in both methods and estimate the inner
expectation in Eq.~\ref{eq:isbbvi_high_dim} with only one sample. For
the outer expectation, we use $8$ samples to estimate the gradient
itself and $8$ separate samples to estimate the optimal coefficient
$a_n$ for the control variates. For \gls{BBVI}, we also doubled the
number of samples to $16+16$; this is marked as ``\gls{BBVI}
($\times 2$)'' in the plots.

For \gls{OBBVI}, we study both a single proposal and a mixture
proposal with two components, respectively labeled as
``\gls{OBBVI} (single proposal)'' and ``\gls{OBBVI} (mixture).''
For the latter, we fix the dispersion coefficients $\tau_{n1}=1$
for all hidden variables and run stochastic gradient descent steps
for $\tau_{n2}$.

At each iteration (and for each method) we evaluate several
quantities: the \gls{ELBO}, the averaged sample variance of the
estimator of the gradient, and a model-specific performance metric on
the test set. The estimation of the \gls{ELBO} is based on a single
sample of the variational distribution $q(\bz;\blambda)$ for all
methods.
For the \gls{GNTS} model, we compute the average log-likelihood (up to
a constant term) on the test set, which is generated with one
additional time instant in all sequences. For the Poisson \gls{DEF},
we compute the average held-out perplexity, \begin{equation}
  \exp\left(\frac{-\sum_{d}\sum_{w\in \textrm{doc}(d)} \log p(w\g \#\textrm{held out in } d)}{\# \textrm{held out words}}\right),
\end{equation}
where the held-out data contains $25\%$ randomly selected words of
all documents.

\parhead{Experimental setup.} For each method, we initialize the
variational parameters to the same point and run each algorithm with a
fixed computational budget (of CPU time).

We use AdaGrad \citep{Duchi2011} for the learning rate. We set the
parameter $\eta$ in Eq.~\ref{eq:fullalg_rho_t} to $\eta=0.5$ for the
\gls{GNTS} model and $\eta=1$ for the Poisson \gls{DEF}. When
optimizing the \gls{OBBVI} dispersion coefficients $\tau_n$, we
take steps of length $0.1$ in the direction of the (negative) gradient.
We initialize the dispersion coefficients as $\tau_n=2$ for the single
proposal and $\tau_{n2}=3$ for the two-component mixture.

We parameterize the normal distribution in terms of its mean and
variance, the gamma in terms of its shape and mean, and the Poisson in
terms of its mean parameter. In order to avoid constrained
optimization, we apply the transformation
$\lambda^\prime=\log(\exp(\lambda)-1)$ to those variational parameters
that are constrained to be positive and take stochastic gradient
steps with respect to $\lambda^\prime$.

\parhead{Overdispersed exponential families.}
For a fixed dispersion coefficient $\tau$, the overdispersed
exponential family of the Gaussian distribution with mean $\mu$ and
variance $\sigma^2$ is a Gaussian distribution with mean $\mu$ and
variance $\tau\sigma^2$. The overdispersed gamma distribution with
shape $s$ and rate $r$ is given by a new gamma distribution with shape
$\frac{s+\tau-1}{\tau}$ and rate $\frac{r}{\tau}$. The overdispersed
$\textrm{Poisson}(\lambda)$ distribution is a
$\textrm{Poisson}(\lambda^{1/\tau})$ distribution.

\subsection{Results}
\label{sec:results}

\begin{figure}[t]
  \centering
  \subfloat[Averaged sample variance of the estimator.\label{fig:gamNormTS_varall_vs_t}]{\includegraphics[width=0.9\columnwidth]{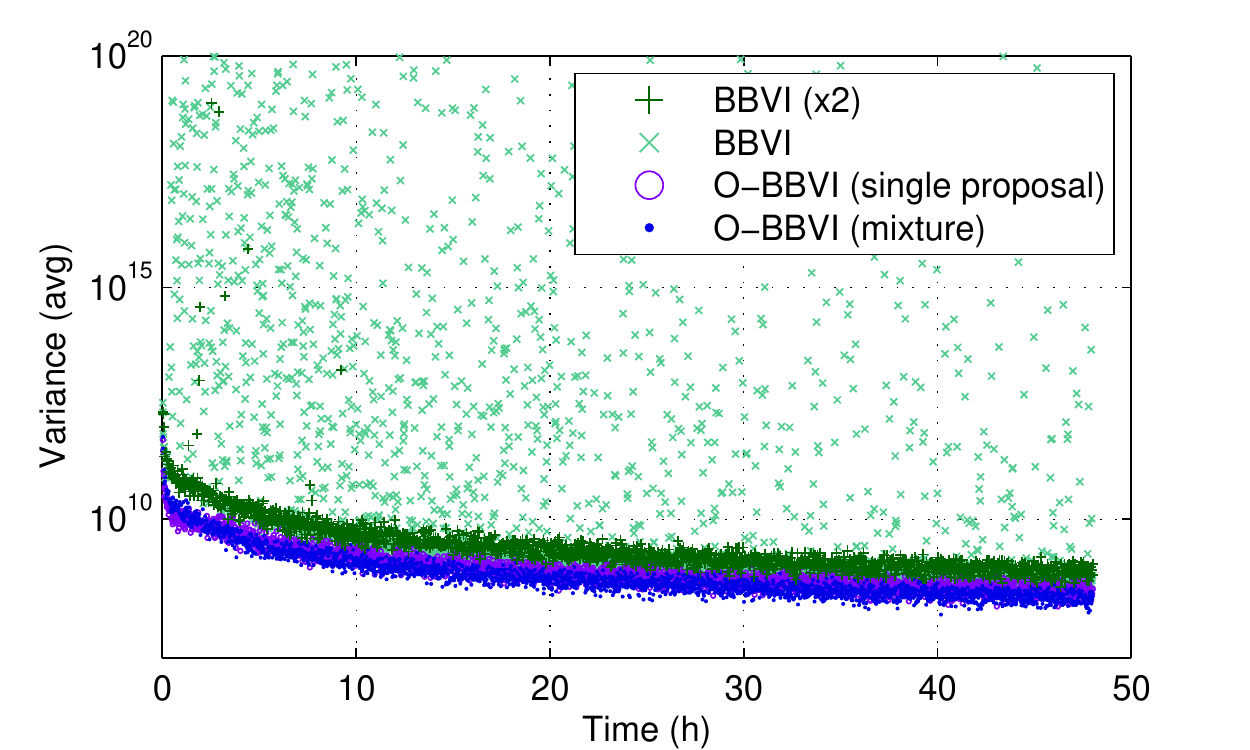}}\\ \vspace*{-7pt}
  \subfloat[Traceplot of the \gls{ELBO}.\label{fig:gamNormTS_elbo_vs_t}]{\includegraphics[width=0.9\columnwidth]{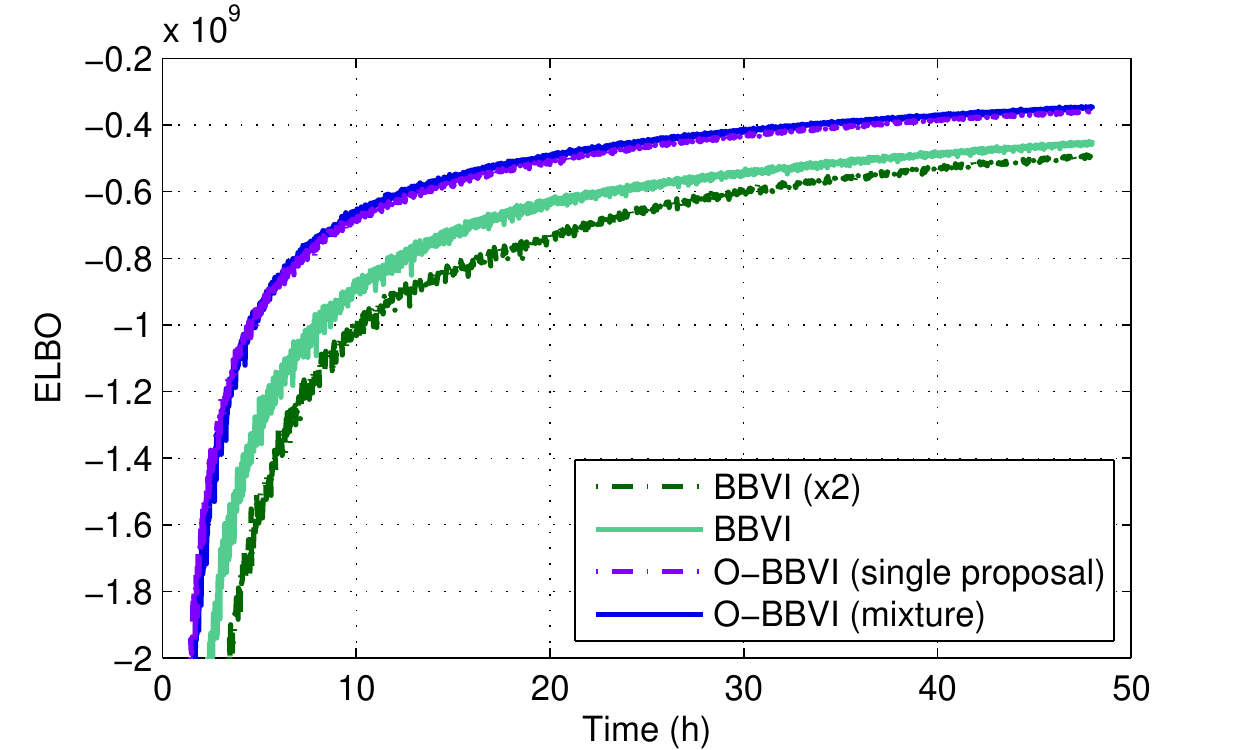}}\\ \vspace*{-12pt}
  \subfloat[Predictive performance (higher is better).\label{fig:gamNormTS_err_vs_t}]{\includegraphics[width=0.9\columnwidth]{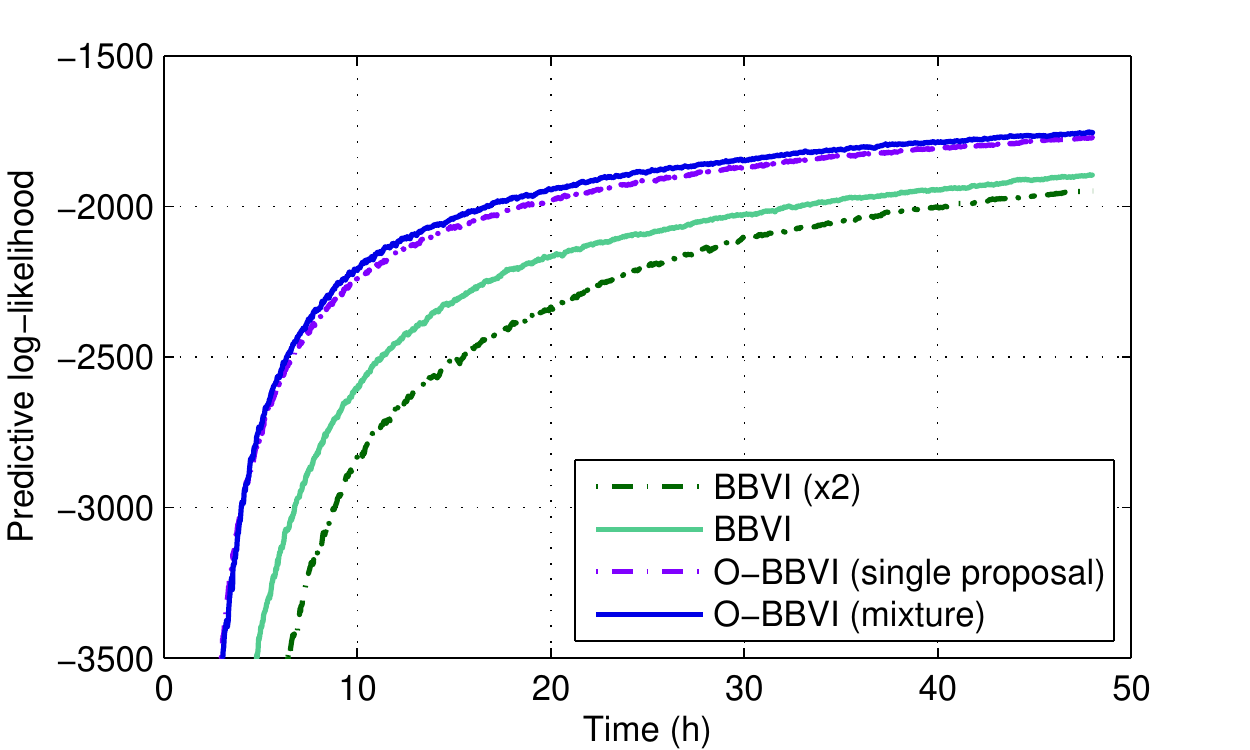}}\\ \vspace*{-4pt}
  \caption{Results for the \gls{GNTS} model.\label{fig:gamNormTS}} \vspace*{-8pt}
\end{figure}

\begin{figure}[t]
  \centering
  \subfloat[Averaged sample variance of the estimator.\label{fig:poissonDEF_varall_vs_t}]{\includegraphics[width=0.9\columnwidth]{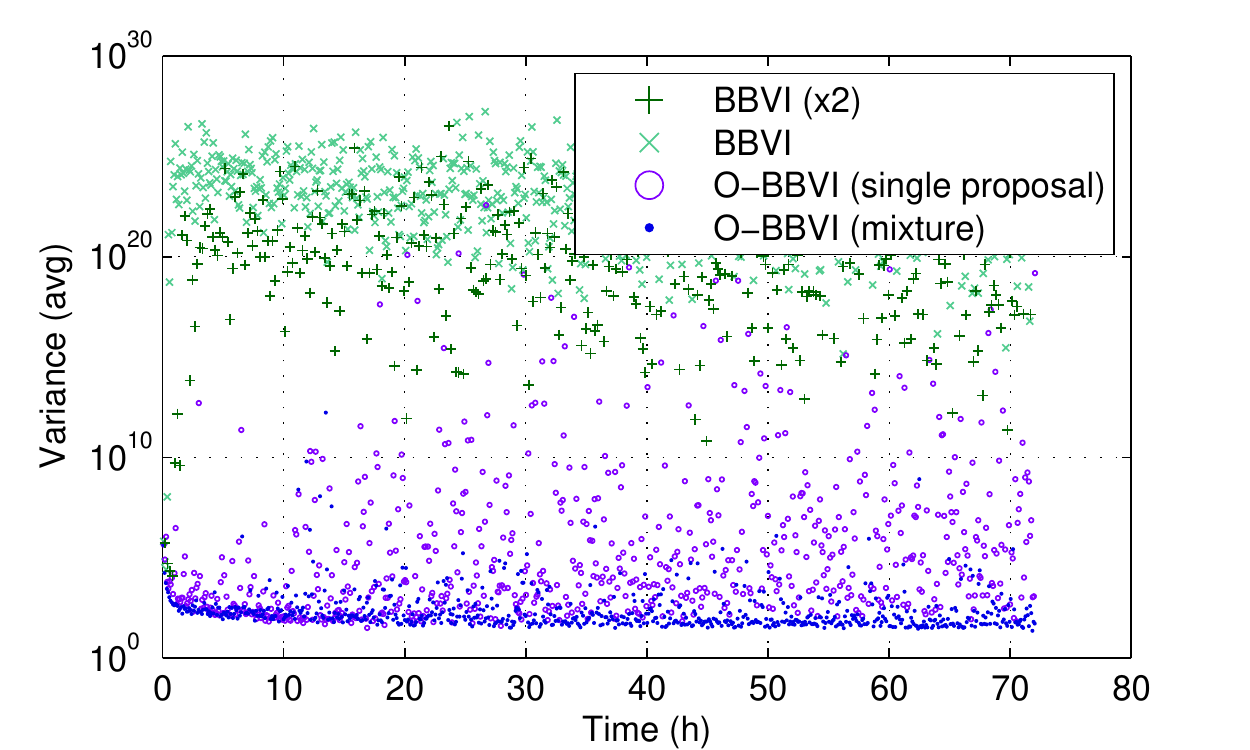}}\\ \vspace*{-7pt}
  \subfloat[Traceplot of the \gls{ELBO}.\label{fig:poissonDEF_elbo_vs_t}]{\includegraphics[width=0.9\columnwidth]{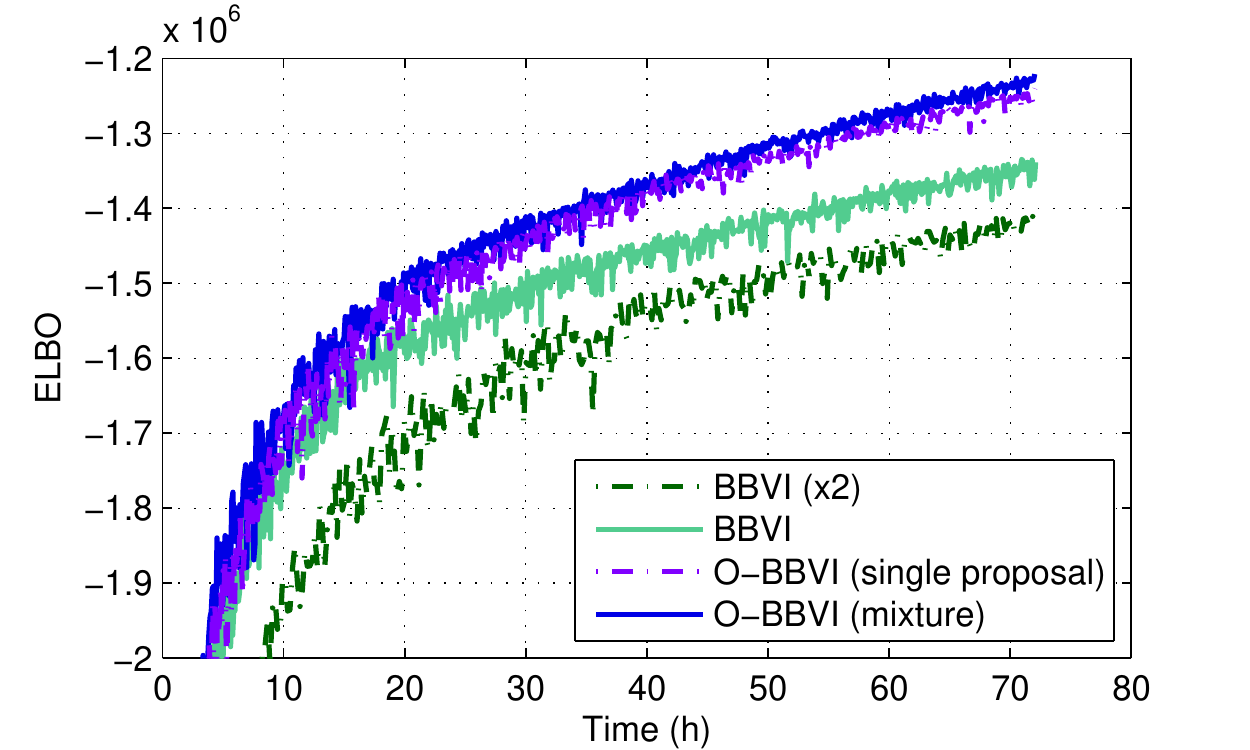}}\\ \vspace*{-12pt}
  \subfloat[Predictive performance (lower is better).\label{fig:poissonDEF_err_vs_t}]{\includegraphics[width=0.9\columnwidth]{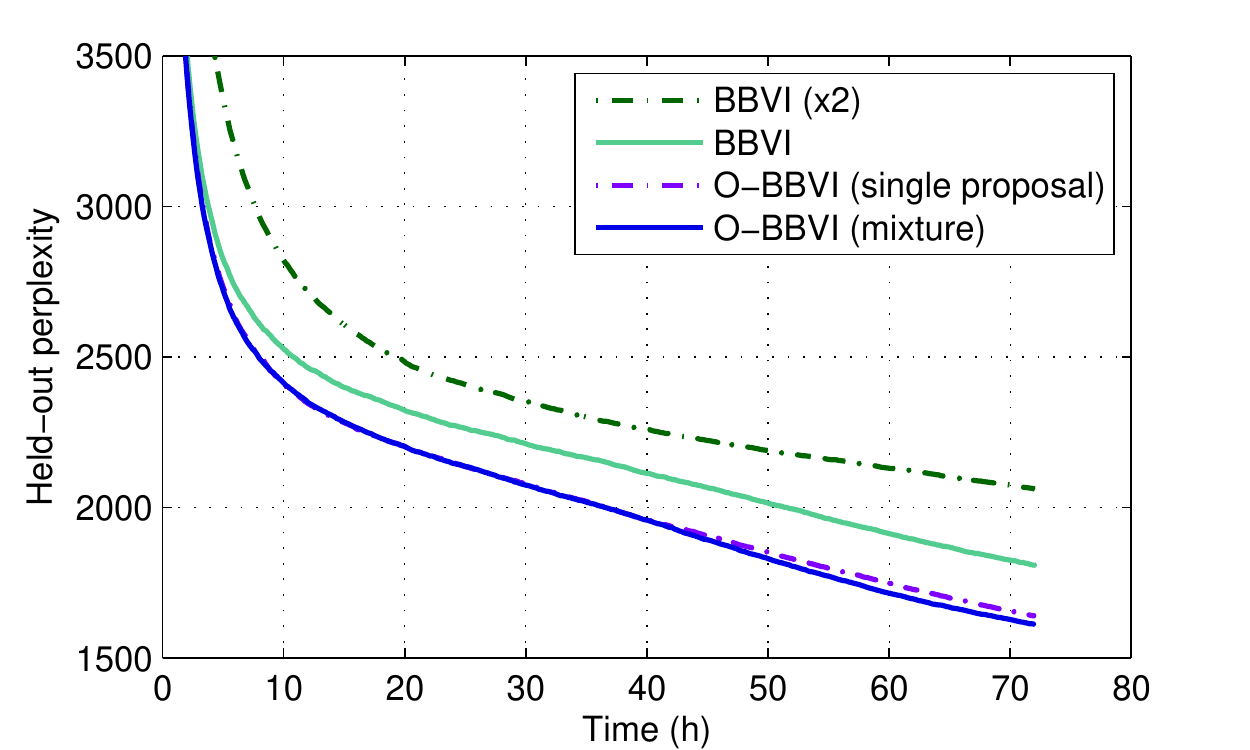}}\\ \vspace*{-4pt}
  \caption{Results for the Poisson \gls{DEF} model.\label{fig:poissonDEF}} \vspace*{-8pt}
\end{figure}

Figures~\ref{fig:gamNormTS} and \ref{fig:poissonDEF} show the
evolution of the \gls{ELBO}, the predictive performance, and the
average sample variance of the estimator for both models and all
methods. We plot these metrics as a function of running time, and
each method is run with the same computational budget.

For the \gls{GNTS} model, Figure~\ref{fig:gamNormTS_varall_vs_t} shows
that the variance of \gls{OBBVI} is significantly lower than
\gls{BBVI} and \gls{BBVI} with twice the number of samples.
Additionally, Figures~\ref{fig:gamNormTS_elbo_vs_t} and
\ref{fig:gamNormTS_err_vs_t} show that \gls{OBBVI} outperforms vanilla
\gls{BBVI} in terms of both \gls{ELBO} and held-out likelihood.
According to these figures, using a single or mixture proposal does
not seem to significantly affect performance.

The results on the Poisson \gls{DEF} are similar
(Figure~\ref{fig:poissonDEF}). Figure~\ref{fig:poissonDEF_varall_vs_t}
shows the average sample variance of the estimator; again, \gls{OBBVI}
outperforms both \gls{BBVI} algorithms.
Figures~\ref{fig:poissonDEF_elbo_vs_t} and
\ref{fig:poissonDEF_err_vs_t} show the evolution of the \gls{ELBO} and
the held-out perplexity, respectively, where \gls{OBBVI} also outperforms
\gls{BBVI}. Here, the two-component mixture proposal performs slightly better than
the single proposal. This is consistent with
Figure~\ref{fig:poissonDEF_varall_vs_t}, which indicates that the
mixture proposal gives more stable estimates than the single proposal.

Finally, for the \gls{GNTS} model only, we also apply the local
expectations algorithm of \citet{Titsias2015}, which relies on exact
or numerical integration to reduce the variance of the estimator. We
form noisy gradients using numerical quadratures for the Gaussian random
variables and standard \gls{BBVI} for the gamma variables (these results
are not plotted in the paper). We found that local expectations
accurately approximate the gradient for the Gaussian distributions.
It converges slightly faster at the beginning of the run, although
\gls{OBBVI} quickly reaches the same performance. (We conjecture
that it is faster because the local expectations algorithm does not
require the use of control variates. This saves evaluations of
the log-joint probability of the model and thus it can run more
iterations in the same period of time.)

However, we emphasize that \gls{OBBVI} is a more general
algorithm than local expectations. The local expectations of
\citet{Titsias2015} are only available for discrete distributions with
finite support and for continuous distributions for which numerical
quadratures are accurate (such as Gaussian distributions). They fail
to approximate the expectations for other exponential family
distributions (e.g., gamma,\footnote{Although
  the univariate gamma distribution is amenable to numerical
  integration, we have found that the approximation of the
  expectations are not accurate when the shape parameter of the gamma
  distribution is below $1$, due to the singularity at $0$.} Poisson,
and others). For example, they cannot handle the Poisson \gls{DEF}.

